\documentclass[10pt,twocolumn,letterpaper]{article}

\usepackage{wacv}
\usepackage[pagebackref,breaklinks,colorlinks]{hyperref}
\usepackage[accsupp]{axessibility}
\usepackage[capitalize]{cleveref}
\crefname{section}{Sec.}{Secs.}
\Crefname{section}{Section}{Sections}
\Crefname{table}{Table}{Tables}
\crefname{table}{Tab.}{Tabs.}

\def\wacvPaperID{8} % *** Enter the WACV Paper ID here
\def\confName{WACV}
\def\confYear{2025}

\usepackage[utf8]{inputenc} % allow utf-8 input
\usepackage[T1]{fontenc}    % use 8-bit T1 fonts
\usepackage{url}            % simple URL typesetting
\usepackage{booktabs}       % professional-quality tables
\usepackage{amsfonts}       % blackboard math symbols
\usepackage{nicefrac}       % compact symbols for 1/2, etc.
\usepackage{microtype}      % microtypography
\usepackage{xcolor}         % colors
\usepackage{amssymb}
\usepackage{graphicx}
\usepackage{caption}
\usepackage[moderate]{savetrees}

\title{EarthView: A Large Scale Remote Sensing Dataset for Self-Supervision }

\begin{document}

\author{%
Diego Velazquez$^{1,}$$^4$, 
Pau Rodriguez L\'opez$^{1,}$$^5$, 
Sergio Alonso$^2$, 
Josep M. Gonfaus$^2$,
Jordi Gonzalez$^1$, \\
Gerardo Richarte$^2$, 
Javier Marin$^2$, 
Yoshua Bengio$^3$, 
Alexandre Lacoste$^4$ \\
$^1$\small Computer Vision Center 
$^2$Satellogic 
$^3$\small Mila, Universit\'e de Montr\'eal
$^4$ServiceNow Research
$^5$Apple Research 
%\small \{diegovd0296, poal\}@gmail.com, \{pau.rodriguez@apple.com\} \{sergio.alonso, josep.gonfaus,gera, javier.marin\}@satellogic.com\\ 
%\small \{poal, pau.rodriguez\}@cvc.uab.es yoshua.bengio@umontreal.ca, alexandre.lacoste@servicenow.com
}

\maketitle

\begin{abstract}
This paper presents EarthView, a comprehensive dataset specifically designed for self-supervision on remote sensing data, intended to enhance deep learning applications on Earth monitoring tasks. The dataset spans 15 tera pixels of global remote-sensing data, combining imagery from a diverse range of sources, including NEON, Sentinel, and a novel release of 1m spatial resolution data from Satellogic. Our dataset provides a wide spectrum of image data with varying resolutions, harnessed from different sensors and organized coherently into an accessible HuggingFace dataset\footnote{Available at \url{https://huggingface.co/datasets/satellogic/EarthView}} in parquet format. This data spans five years, from 2017 to 2022. Accompanying the dataset, we introduce EarthMAE, a tailored Masked Autoencoder, developed to tackle the distinct challenges of remote sensing data. Trained in a self-supervised fashion, EarthMAE effectively processes different data modalities such as hyperspectral, multispectral, topographical data, segmentation maps, and temporal structure. This model helps us show that pre-training on Satellogic data improves performance on downstream tasks.  While there is still a gap to fill in MAE for heterogeneous data, we regard this innovative combination of an expansive, diverse dataset and a versatile model adapted for self-supervised learning as a stride forward in deep learning for Earth monitoring.
\end{abstract}

\section{Introduction} 

The instrumental role of Earth monitoring in navigating and confronting the escalating challenges of climate change, natural disasters, and environmental issues cannot be overstated~\cite{rolnick2022tackling}. It is also increasingly playing a central role in agriculture \cite{khanal2020remote} and city planning \cite{netzband2007applied}. Traditional vision models have already provided significant value across myriad applications \cite{radford2021learning,rodriguez2017deep,laradji2020counting}, and the emergence of foundation models \cite{openai2023gpt4, oquab2023dinov2, lacoste2021foundation} promises to bring Earth monitoring to new horizons. 

Today's large vision models, while proficient in detection \cite{ren2015faster} and image segmentation \cite{fang2023eva}, are largely designed for RGB images from a first-person perspective. In contrast, remote sensing data offers unique properties: a \textbf{sky view} with a wide range of spatial resolutions, \textbf{multi-modality} including multispectral, radar, hyperspectral, point clouds, and elevation maps, and \textbf{temporality} from revisiting the same locations multiple times. Essentially, data with geographic coordinates can be matched with other data from the same location, providing an ever-expanding source of structured data suitable for large-scale self-supervised learning algorithms.

However, despite the surfeit of data that exceeds our current algorithmic processing capabilities, a significant portion of it remains out of reach, locked behind expensive paywalls. Free data sources such as Sentinel-1 and Sentinel-2, while useful, come with limitations: i) \textbf{low spatial resolution} of 10m of ground sample distance (GSD) and ii) \textbf{download difficulties} due to bandwidth throttling on Google Earth Engine and the cost associated with AWS for large-scale downloads.

To bridge this gap, we have teamed up with Satellogic and NEON to release a 15 tera pixels comprehensive, large-scale dataset designed specifically for self-supervised learning of extensive Earth monitoring algorithms. The dataset is available at Hugging Face and is conveniently partitioned to allow working on subsets of the data. This robust dataset comprises structured data derived from three distinct sources:

\begin{itemize}
\item Satellogic: Provides RGB and near-infrared data at 1m GSD, with temporal revisits and planet coverage.
\item NEON: Provides 369 bands of hyperspectral data at 1m GSD, complemented by RGB data at 0.1m GSD and elevation data at 1m GSD across various US forests.
\item Sentinel: We gathered a large structured subset of Sentinel-1 and 2, combining multispectral,  synthetic aperture radar (SAR), and temporality.
\end{itemize}

The key contributions of this work are twofold:

\begin{itemize}
    \item The introduction of a large-scale, multi-modal dataset tailored specifically for self-supervised learning in Earth monitoring.
    \item The development of a large masked auto-encoder trained with various self-supervision schema, demonstrating high performance across various Earth monitoring tasks.
\end{itemize}

\section{Related Work}

\subsection{Datasets for training}

%e.g. satlas, planet (TODO(allac) find it). 

The success of large-scale deep learning models has triggered research on larger datasets that can fit the capacity of current systems.
\cite{sumbul2019bigEarthnet} introduced BigEarthNet, a large-scale benchmark archive for remote sensing image understanding. This dataset consists of 590,326 Sentinel-2 image patches, annotated with multiple land-cover classes. The annotations were provided by the CORINE Land Cover database, and the dataset was significantly larger than existing archives in remote sensing. The authors demonstrated that training models on BigEarthNet improved accuracy compared to pre-training on ImageNet, indicating its potential for advancing operational remote sensing applications. \cite{qi2020mlrsnet} addressed the need for multi-label annotated datasets in remote sensing for semantic scene understanding. They developed MLRSNet, a multi-label high spatial resolution remote sensing dataset containing 109,161 samples within 46 scene categories. Each image in MLRSNet has at least one of 60 predefined labels, enabling training deep learning models for multi-label tasks such as scene classification and image retrieval. The authors highlighted the importance of MLRSNet as a benchmark dataset and its complementary nature to existing datasets like ImageNet.
 \cite{tong2023enabling} presented the Five-Billion-Pixels dataset, aiming to enable country-scale land cover mapping with meter-resolution satellite imagery. The dataset comprises more than 5 billion labeled pixels from 150 high-resolution Gaofen-2 satellite images. They proposed a deep-learning-based unsupervised domain adaptation approach to transfer classification models trained on labeled data to unlabeled data for large-scale land cover mapping. The experiments demonstrated promising results across different sensors and geographical regions, showcasing the potential of the dataset and proposed approach. In a concurrent work, \cite{bastani2022satlas} introduced Satlas, a large-scale dataset for remote sensing image understanding. Satlas is comprehensive in terms of both breadth and scale, containing 302 million labels across 137 categories over a cumulative of 17 trillion pixels. The authors evaluated multiple baselines and a proposed method on Satlas. Pre-training on Satlas significantly improved performance on downstream tasks compared to ImageNet and other baselines. Lastly, the Umbra Open Data Program \cite{umbra_open_data_2025} features over twenty diverse time-series locations that are updated frequently, allowing users to experiment with SAR’s capabilities.

While the previous benchmarks constitute a significant step in data availability for remote sensing, they are typically limited by the cost of obtaining labels. This has motivated the construction of unlabeled datasets that can leverage uncurated data from many different sources. For example, \cite{manas2021seasonal} proposed to leverage unlabeled data with Seasonal Contrast (SeCo). They collected a dataset of Sentinel-2 patches without human supervision, consisting of 1 million multispectral image patches from approximately 200,000 locations worldwide. By capturing seasonal changes with images from different dates, they aimed to enhance the training of models for remote sensing tasks. In a similar fashion, \cite{stewart2024ssl4eo} proposed SSL4EO-L, consisting of 5 million unlabeled image patches from Landsat across 250,000 locations and multiple seasons. While SeCo and SSL4EO-L focused on uniformly covering most of the inhabited regions of Earth from a single data source,  \cite{bhugra2022rapidai4eo} focused on densely covering Europe with multiple data sources (Copernicus, Sentinel-2, and Planet) over space and time (500,000 locations in Europe with daily readings for a year) while \cite{scheibenreif2023masked} used EnMAP as their single source. In this work, we combine multiple data sources at different points in time while considering most of the inhabited Earth, resulting in a dataset that we named EarthView. Concretely, EarthView offers a larger and more diverse collection of unlabeled data by combining a high-quality curated selection from multiple data sources (Sentinel, NEON, and Satellogic), achieving a larger scale and variety than previous works (over 15 trillion pixels, with temporal revisits, and from 60 to 0.1m resolution). We share EarthView in a highly accessible format and available through Hugging Face, which enables easy integration into research projects. These qualities make our dataset a valuable resource for exploring uncharted patterns and structures in an unsupervised learning setting.

\subsection{Learning from unlabelled data}

\begin{figure}[t!]
     \centering
     \includegraphics[width=\columnwidth]{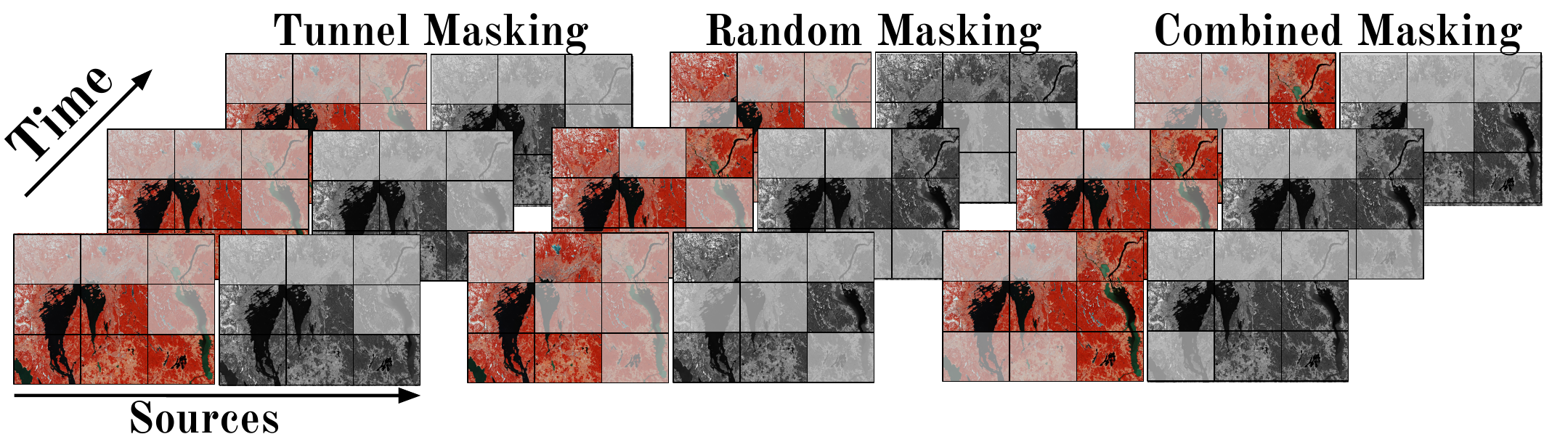}
     \caption{Different masking schemes explored in our work. Random masking, masks random patches across sources/time while tube masking masks the same patches. Combined masking combines both by first masking some patches consistently across sources/time and then randomly masking a subset of the remaining ones.}
    \label{fig:masking}
\end{figure}

Multi-view self-supervised learning methods have played a crucial role in building large models with remote sensing data~\cite{manas2021seasonal,bayrooti2023multispectral, wanyan2023dino}. In addition to multi-view SSL, reconstruction-based SSL with MAEs~\cite{he2022masked} has also been explored in the context of remote sensing. \cite{wang2023remote} introduced MIM, using masked image modeling for remote sensing scene classification. SatMAE \cite{cong2022satmae} introduced a pre-training framework leveraging temporal and multispectral satellite imagery, encoding groups of bands independently with a spectral positional encoding. Scale-MAE and SatMAE++ additionally leverage information from multiple scales~\cite{reed2022scale,noman2024rethinking}. SpectralMAE \cite{zhu2023spectralmae} and DOFA \cite{xiong2024neural}, focused on the reconstruction of arbitrary combinations of bands and data sources. Given the simplicity and versatility of MAE-based approaches to handling multiple data sources, we choose the SpectralMAE model class to experiment with the EarthView dataset introduced in this work. Concretely, we generalize SatMAE and SpectralMAE by combining multiple masking strategies, i.e. we combine masking all bands given a random position in an image with randomly masking individual bands in random positions (see \cref{fig:masking}). In experiments, we find that this strategy is effective for learning from heterogeneous data sources like the proposed EarthView data. 

%Additionally, supervised transfer learning has also been explored in the context of remote sensing. \cite{fuller2022transfer} pre-trained SatViT-V2 on a large satellite-derived dataset, demonstrating improved performance on both in-distribution and out-of-distribution data. \cite{bastani2022satlas} showed that pre-training a SwinV2 \cite{liu2022swin} architecture on Satlas substantially boosted downstream task performance.

% MAE-based approaches have been widely explored for this purpose \cite{he2022masked}. 
 
% On the other hand, 

% Overall, these works collectively demonstrate the significant progress made in self-supervised learning methods for remote sensing with MAEs and transformers particularly relevant for encoding hyperspectral data with dedicated positional, temporal and spectral encodings. 

% PROMPT FOR PAU

% - We include Satellogic data, being made public for the first time
% - We present a curated version of previously available data (sentinel, neon)
% - I guess we could claim that we try combined masking (not tried before) and it actually improves performance on some benchmarks. (sounds simple but it was a pain in the ass to code)

% Regarding datasets, we have adopted a different approach from existing benchmarks by focusing on unlabeled datasets that exploit uncurated data from a variety of sources. Unlike previous works that primarily center on large, labeled datasets, our strategy seeks to leverage the vast amount of available unlabeled data, making it more scalable and cost-effective.

\section{Dataset}

This work introduces an extensive dataset tailored for self-supervision on Earth monitoring data. It is based on the assumption that structure in the data brings an essential signal to self-supervised algorithms for finding high-level semantic representation that can make sense of the data. To this end, we combine spectral, synthetic-aperture radar (SAR), temporal, and spatial structures in a large-scale dataset composed of different sources and multiple spatial resolutions.
% for advancing the application of deep learning methodologies within the realm of climatological research. Encompassing roughly 12 terabytes of globally dispersed satellite-derived data, this dataset functions as a robust tool for facilitating the development and refinement of potent machine-learning algorithms.
\begin{figure}
     \centering
     \includegraphics[width=\columnwidth]{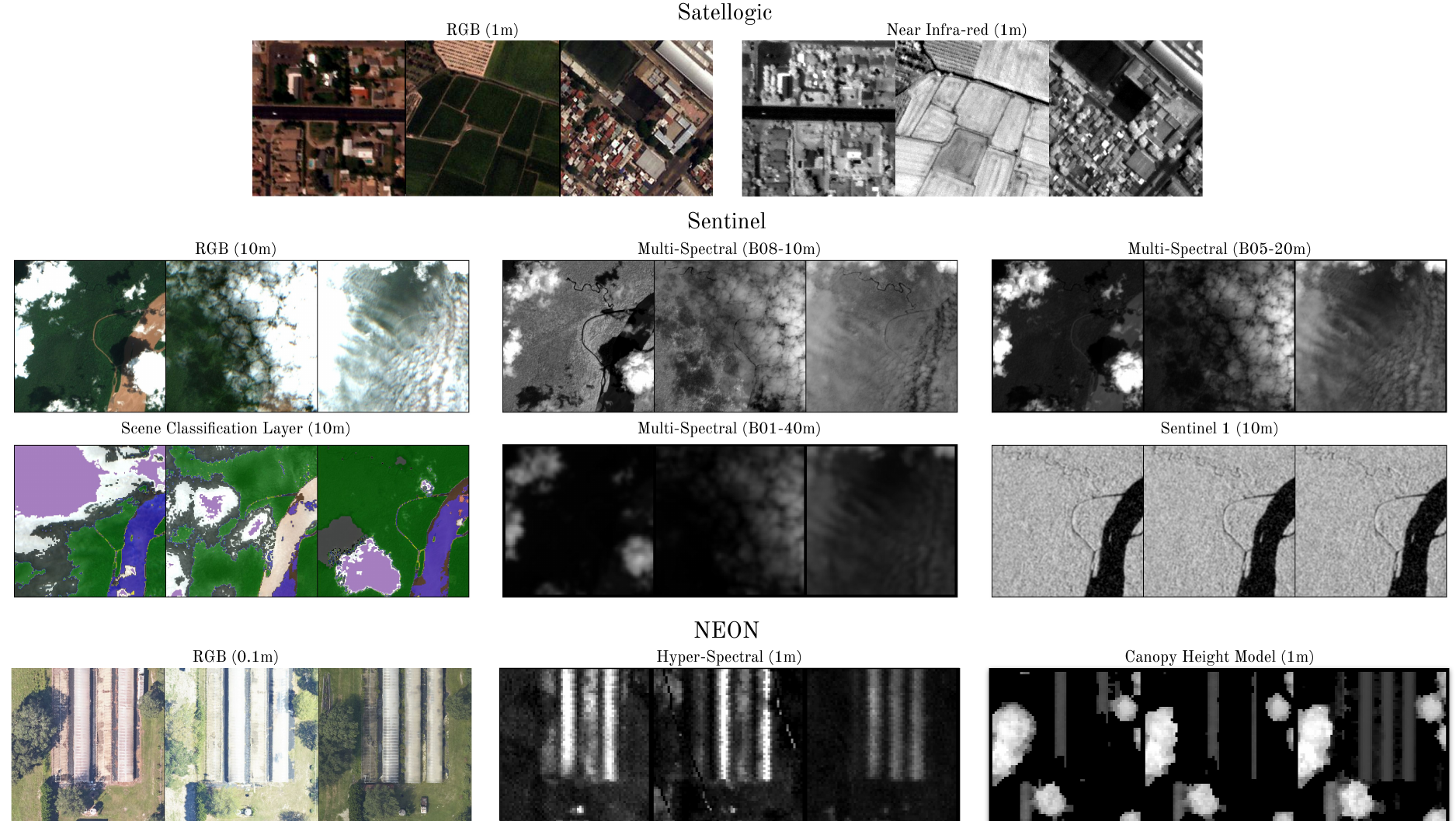}
     \caption{Samples from the dataset}
    \label{fig:samples}
\end{figure}
% \subsection{Data Acquisition and Composition}
The data gathered for this project is drawn from a triad of distinct sources, namely Satellogic, Sentinel and NEON (\cref{fig:samples}). Each of these contributes unique facets and dimensionalities to the integrated dataset.

\subsection{Satellogic}

\paragraph{Vision} Satellogic is a provider of high-resolution remote sensing imagery, whose main mission consists of democratizing access to Earth Observation Data. To date, Satellogic's fleet consists of 24 active satellites, having successfully put over 50 satellites in orbit in the past years. These satellites provide high-resolution imagery that ranges from 50cm to 1m. 

Building multimodal foundation models is paramount to further advance Earth Observation applications. Thanks to these models, the EO community can easily construct on top of the new solutions without the need to collect large amounts of data. Multimodal foundation models, such as vision-language large models, to be effective require to be trained on vast amounts of data. More importantly, in this specific case, data should cover a large variety of regions over the planet to have the most accurate representation of Earth. To this end, Satellogic offers to release a portion of its second half of 2022, ranging from the 1st of July to the 30th of December. 

\paragraph{Accessibility} High-resolution data covering different regions of the world is not freely accessible. To truly push the community towards building open foundation models, Satellogic releases its dataset under the license \href{https://creativecommons.org/licenses/by/4.0/}{CC BY 4.0}. To our knowledge, this release represents one of the largest contribution to the EO community ever made by a private company. Moreover, this is the first one that includes the visual and top-of-atmosphere products consisting both of four bands (red, green, blue and near-infrared) at a 1-meter resolution. Satellogic also offers metadata such as the off-nadir angle, sun elevation, azimuth angle, or timestamp. The Satellogic dataset stands out as unique compared to other publicly available datasets for several reasons: 
\begin{itemize}
    \item It covers diverse regions of the Earth, unlike \textbf{NAIP}~\cite{naip}, \textbf{LINZ}~\cite{linz}, or \textbf{NEON}
    \item It offers high-resolution imagery, surpassing traditional open data sources like \textbf{Sentinel-2} and \textbf{Landsat 8 and 9}
    \item It allows for commercial use, as opposed to datasets such as \textbf{DOTA}~\cite{dota}, \textbf{UC Merced Land-Use}~\cite{merced} or \textbf{xView}~\cite{xview}
    \item It includes rich metadata and geolocation
    \item Its size is multiple orders of magnitude larger than most of the publicly available datasets, \textit{e.g.} \textbf{FAIR1M}~\cite{fair1m}, \textbf{DIOR}~\cite{dior}, \textbf{NWPU-RESISC-45}~\cite{nwpu} and \textbf{Floodnet}~\cite{floodnet}
\end{itemize}

\paragraph{Sensor} Imagery is acquired at 1m GSD from space, at different off-nadir angles, over four bands (blue, green, red and near-infrared). 
\vspace{-2.5mm}
\paragraph{Spatial Distribution} The acquisition of imagery during the 2nd half of 2022 was performed on demand by multiple customers. As can be seen in Fig.~\ref{fig:coverage}, the Satellogic dataset covers different regions over all the continents. To build these regions, we make use of 3,758 captures, where all these captures have a percentage of clouds below 30\%. Out of these captures, we create unique patches. In particular, for each capture, we start at the top-left position and crop non-overlapping windows of 384 $\times$ 384 pixels. %For areas where we have multiple captures, depending on how they overlap, it can result in regions that overlap with other ones. 
% While this was initially a non-desired behaviour, we kept them as a sort of data augmentation. 
The dataset comprises a total of 2,967,663 patches, summing up $437,682$ km$^2$. If we discard overlapping regions\footnote{An artifact of our patching algorithm}, the dataset covers a 10\% smaller area ($396,280$ km$^2$). Most of these overlapping regions have less than 50\% redundancy and are comparable to data augmentation.
%In total, there are $549,089$ unique regions that overlap with each other, resulting in $2,649,018$ overlapping pairs. Out of these pairs, $2,252,434$ overlap less than $50\%$, and only $7,553$ overlap more than $90\%$.
\vspace{-2.5mm}
\paragraph{Temporality} The resulting set of patches contains a varying number of revisits, depending on how many captures were performed over the same area. These revisits range from 1 to 68, where 986,521 regions have at least two revisits. The almost 3M patches translate to 6,165,992 images including revisits. 

\subsection{Sentinel} 
\paragraph{Accessibility} While Sentinel data is distributed under a Creative Commons license, very large datasets are less accessible. Since the Google Earth Engine (GEE) throttles the download speed, it is prohibitively long to download terabytes of data. 
We thus had to resort to AWS, but since \emph{requester pays} the bandwidth, it still required a large budget just to collect this data.\footnote{AWS stores data in large tiles, requiring many downloads for broad coverage despite needing only fractions.}
\vspace{-2.5mm}
\paragraph{Sensors} Sentinel's satellites offer a wide range of sensors. For this project, we focus on SAR from \textbf{Sentinel-1}, and multispectral from \textbf{Sentinel-2}. The other sensors offer a spatial resolution that is too low for our purpose. For SAR, we use the level-1 Ground Range Detected (GRD) product available in AWS. We stack the different polarizations (VH, VV) resulting in two bands, and resample it to 10m resolution. Sentinel-1 images are then saved in uint16 format to reduce their size in bytes. Finally, Sentinel-2 is composed of 13 spectral bands. The main bands, (blue, green, red, near-infrared), have 10m GSD, but due to atmospheric absorption of other wavelengths other bands have 20m and 60m resolution. %For ease of use, we resampled the 60m resolution to 40m resolution.

\paragraph{Spatial Distribution} 
Our aim is to gather a wide range of regions covering the planet, however, we also want to avoid highly redundant patterns such as ocean, desert, and forests. To this end, we gather inspiration from \cite{manas2021seasonal} and collect Sentinel-2 tiles that overlap regions within a 50km radius around the top largest cities in the world. Each footprint area (100km x 100km) is large enough to cover coastal, agricultural and rural regions. We also sample Sentinel-2 tile regions that cover Satellogic data. Since Sentinel-1 does not follow the same grid system as Sentinel-2, we use the collected Sentinel-2 tile footprints to query Sentinel-1 captures, and crop them accordingly. 

For each Sentinel-2 tile footprint, we extract 500 non-overlapping regions of $3,840$ m x $3,840$ m. Out of all possible candidates (over 670) per tile, we select the best ones based on the cloud coverage and the entropy of Sentinel-2's scene classification layer (to increase diversity within patches). We end up collecting over 2,000 Sentinel-2 tiles, resulting in over 1M unique regions. These same regions are used to build the Sentinel-1 collection. Similar to Satellogic, some of the 1M regions do overlap. The 1M regions sum up a total of $15,474,873$ km$^2$, if we take into account the overlapping areas, the coverage area reduces to $15,074,640$ km$^2$. Out of the 1M regions, there are $65,251$ that overlap, covering $388,840$ km$^2$. A large part of these overlapping regions have less than 50\% redundancy.

\paragraph{Temporal distribution}
Temporality also offers an important signal for a model to learn how scenes evolve over time. However, a long sequence could significantly increase the redundancy and size of the dataset. Hence we limit to ten revisits per location, where six are densely sampled over time and the other four are sampled with three-month intervals to ensure coverage of the seasons. %For this, we allow certain flexibility. %(see Fig.\ref{fig:s1s2} right example). 
At the same time, during the selection process, we check multiple time sequences under these same constraints. For each sequence, we extract the percentage of clouds for all the dates and keep it as an indicator of how good the sequence is. Among all the sequences, we select the best candidate. This results in sequences ranging from 2017 to 2022. For Sentinel-1, we did not have access to the same temporality. During the collection process, we choose the closest dates to the ones we have for Sentinel-2% (see Fig.\ref{fig:s1s2})
. When feasible, we also select other dates available within Sentinel-2 range to increase the number of Sentinel-1 revisits.

% \begin{figure}
%      \centering
%      \includegraphics[width=0.48\textwidth]{figures/comparison_s1_s2_temporality_1.pdf}
%      \includegraphics[width=0.48\textwidth]{figures/comparison_s1_s2_temporality_2.pdf}
%      \caption{Sentinel-1 and Sentinel-2 temporal distribution examples of two different regions.}
%     \label{fig:s1s2}
% \end{figure}

\subsection{NEON}
NEON data combines high-resolution RGB, hyperspectral, and lidar data for the study of ecological sites in the United States. 

\begin{figure}
     \centering
     \includegraphics[width=\columnwidth]{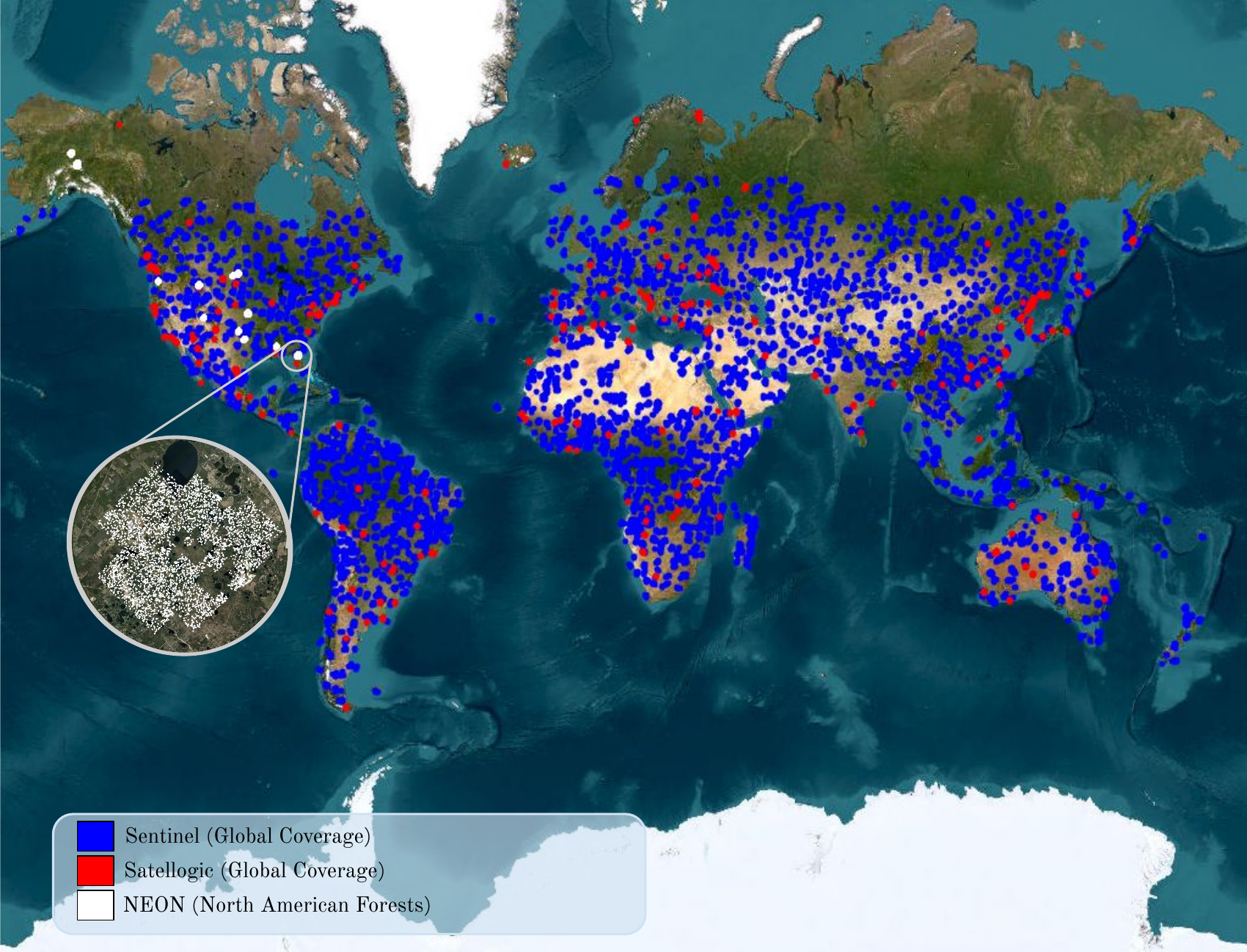}
     \caption{Spatial coverage for each source. Note that a colored area may contain multiple patches.}
    \label{fig:coverage}
\end{figure}

\paragraph{Accessibility} NEON data is redistributed under \href{https://www.neonscience.org/data-samples/data-policies-citation}{CC0 1.0} and accessible on the \href{https://data.neonscience.org/data-products/explore}{NEON data portal}.
\vspace{-2.5mm}
\paragraph{Sensors} NEON offers high-resolution RGB at 0.1m \cite{neon_rgb_provisional} GSD and hyperspectral data \cite{neon_reflectance_provisional} comprised of 426 spectral bands at 1m GSD. It is also accompanied by lidar, which is post-processed to estimate the tree canopy height at 1m GSD \cite{neon_chm_provisional}.
\vspace{-2.5mm}
\paragraph{Spatial Distribution} This incredibly high-resolution data comes with very limited spatial distribution. We have collected data from 12 of the available sites with multiple sub-locations on each of these sites (See \cref{fig:coverage}). Each location covers 64m $\times$ 64m, where depending on the sensor, we have 640 pixels $\times$ 640 pixels or 64 pixels $\times$ 64 pixels. All locations sum up $148.76$ km$^2$. Similar to Satellogic and Sentinel, Neon has some redundancy. The overlapping regions cover less than $5\%$, where most of these overlapping regions have less than $50\%$ redundancy.
\vspace{-2.5mm}
\paragraph{Temporality} This data also offers yearly revisits with some limitations. Most sites contain a maximum of 3 revisits and the exact date was not collected. Nevertheless, we matched all available revisits for each location that we collected.

\begin{table*}[ht]
\caption{Dataset overview}
\label{table:dataset-overview}
\centering
\resizebox{\textwidth}{!}{
\begin{tabular}{r |c c c c c c c c}
 & \textbf{Sensor} & \textbf{\# bands} & \textbf{GSD (m)} & \textbf{Pixel per patch} & \textbf{area (m)} & \textbf{\# revisits} & \textbf{\# patches} & \textbf{\# Giga gray pixels} \\
\hline
\textbf{Satellogic} & RGBN & 4 & 1 & 384 x 384 & 384 x 384 & 1 - 5 & 2967663 & 3636.85 \\
\textbf{Sentinel-1} & SAR & 2 & 10 & 384 x 384 & 3840 x 3840 & 3 - 9 & 1049466 & 1743.61 \\
\textbf{Sentinel-2} & MS & 13 & 10, 20, 60 & 384 x 384 & 3840 x 3840 & 10 & 1049466 & 9086.36 \\
\textbf{NEON-RGB} & RGB & 3 & 0.1 & 640 x 640 & 64 x 64 & 3 & 35501 & 130.87 \\
\textbf{NEON-Hyper} & HS & 369 & 1 & 64 x 64 & 64 x 64 & 3 & 35501 & 160.97 \\
\textbf{NEON-Elev} & Lidar & 1 & 1 & 64 x 64 & 64 x 64 & 3 & 35501 & 0.44 \\
\end{tabular}
}

\end{table*}

\begin{figure}
     \centering
     \includegraphics[width=1.0\columnwidth]{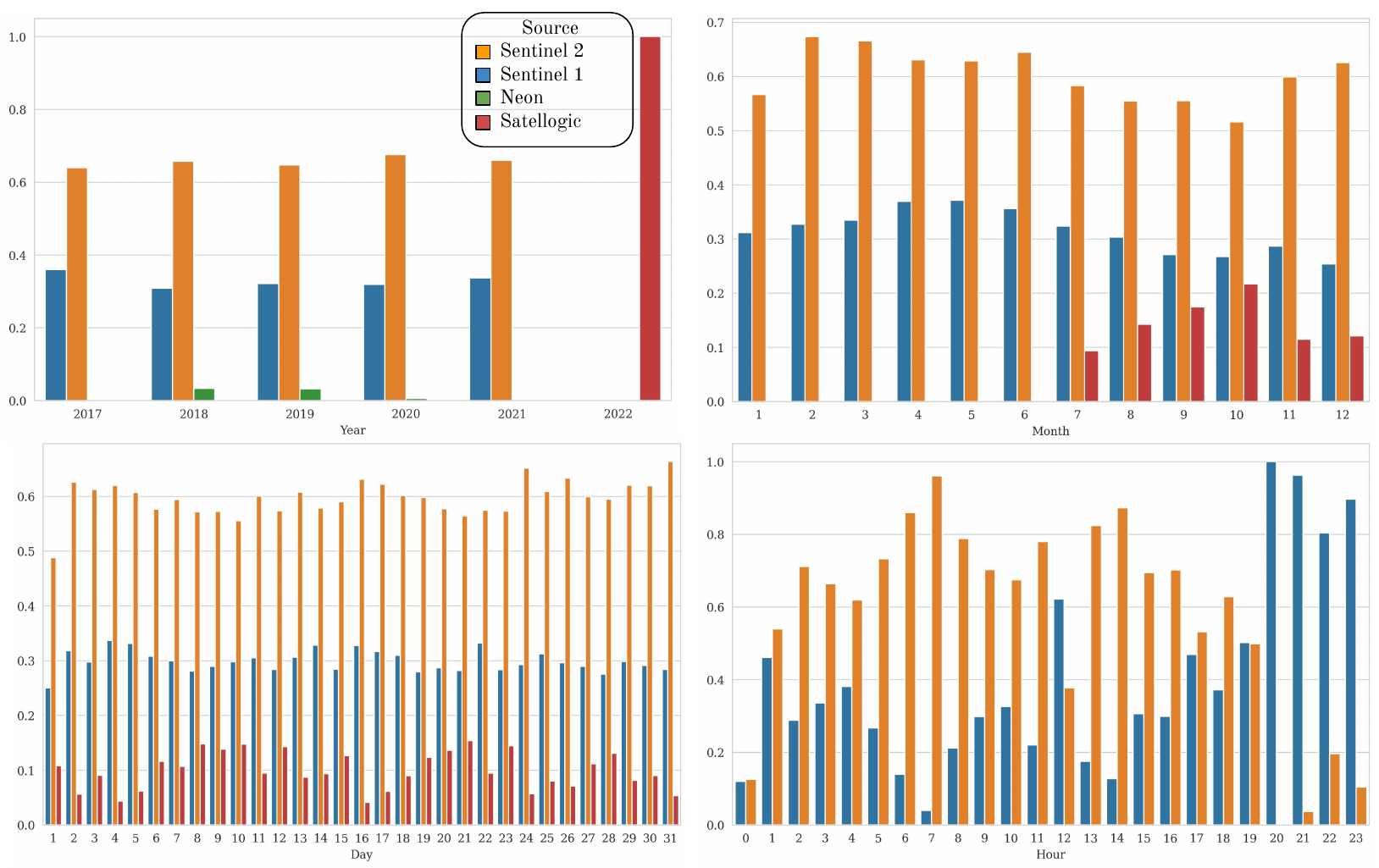}
     \caption{Temporal distribution of the dataset. NEON data only provides the year, and Satellogic data does not contain the time of the day.}
    \label{fig:time}
\end{figure}

\subsection{Hosting and Storage}

\paragraph{Hosting} Through a partnership with HuggingFace, data is made available at their servers.%\url{https://huggingface.co/datasets/satellogic/EarthView}. 
This offers a free download and broad access to the whole community. For other commercial cloud providers, even if hosting were free, downloading 15 TB of data as \emph{requester pays} would cost on the scale of 1000 USD per download. 
\vspace{-2.5mm}
\paragraph{Storage}
For accessibility and to minimize bandwidth, we store the dataset in functional subsets. That is, each of the different sensors can be downloaded separately. Moreover, each subset is split into shards and these can be downloaded independently, this allows the user to download only a set of the data if needed. Each shard contains a series of regions.%\footnote{Regions in the shards are not IID they may be corelated}.
\vspace{-2.5mm}
\paragraph{Format}
Each region is represented as a dictionary containing the sensor data with the RGB data along with the other bands grouped by resolution (e.g., RGB, 10m, 20m, etc ) arrayed in a four-dimensional matrix structure (time, spectral bands, height, width). The dictionary also contains the metadata available for the region (e.g., bounds, timestamps, etc). As metadata, we provide the geo-referenced bounding box and some form of timestamp see (\cref{fig:time}) for all sensors. However other metadata, such as solar angles and incidence angles is only available for Sentinel-2.

\begin{figure*}
     \centering
     \includegraphics[width=0.95\textwidth]{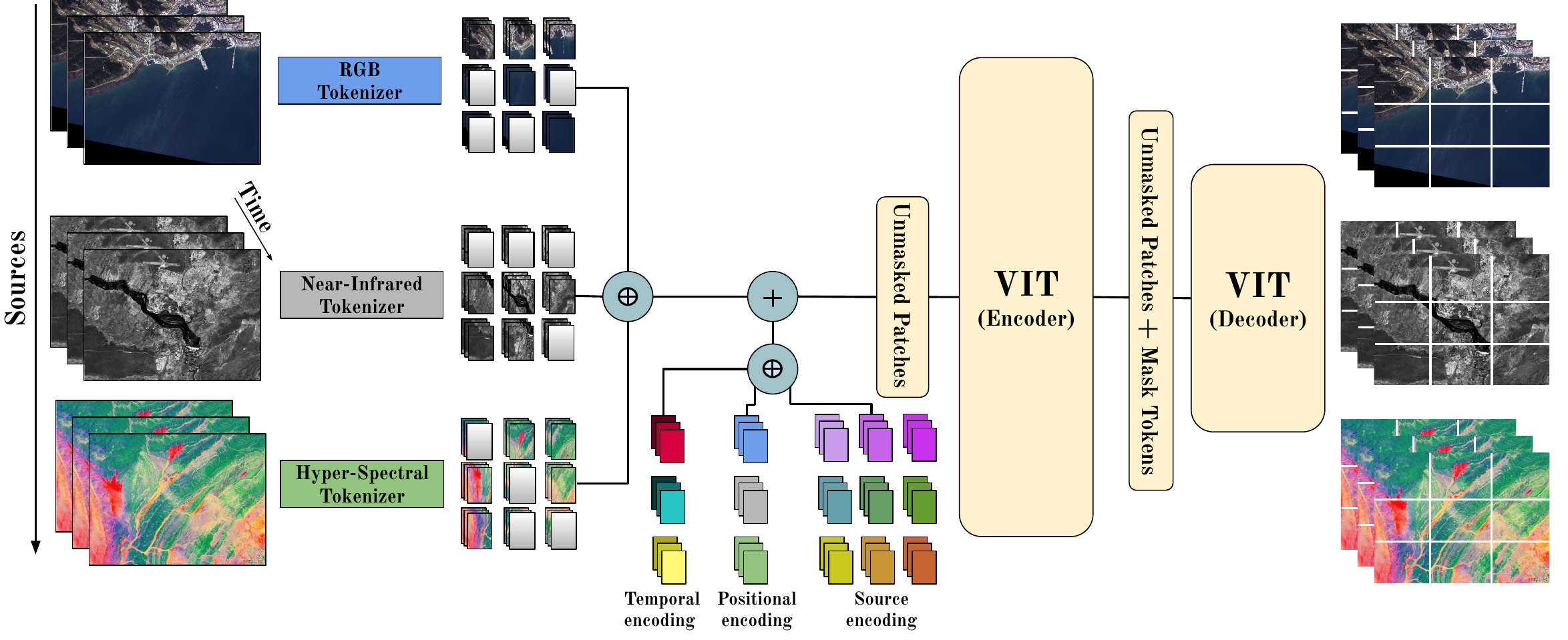}
     \caption{EarthMAE: The model leverages time information and can digest data from an arbitrary number of sources. Each input source is tokenized into a fixed number of patches and then all patches are concatenated. The time, source, and positional encodings are concatenated and added to the patches.}   
    \label{fig:earthmae}
\end{figure*}

\section{Model}

In this work, we leverage a Masked Autoencoder (MAE) \cite{he2021masked}, distinguished by its asymmetrical encoder-decoder architecture. It incorporates an encoder that functions exclusively on a visible subset of patches, and a streamlined decoder, that rebuilds the original image from the latent representation and mask tokens. This model, recognized for its proficiency in self-supervised learning tasks, has been appropriately restructured to manage remote sensing data as described next.

\subsection{EarthMAE}

Our EarthMAE model (\cref{fig:earthmae}) remains faithful to the original architecture, albeit we adjusted the tokenizers and positional encodings to leverage time and different modalities.

\paragraph{Tokenizers} We incorporated a distinct tokenizer (linear transform) for each source, owing to the fact that different sources contain a disparate number of channels. This method offers a more nuanced comprehension of the data, accounting for the varied characteristics associated with different bands and sources. To ensure that all bands and sources produce the same amount of patches we resize all images to $224 \times 224$.
\vspace{-2.5mm}
\paragraph{Encoding} Analogous to positional encodings, we introduce source and temporal encodings to capture the unique characteristics of the data. The source encoding is generated by embedding the source labels into a vector space, allowing the model to differentiate between various data sources such as multispectral, RGB, and hyperspectral images.

For the temporal encoding, we leverage the timestamp metadata provided in our dataset. We decompose each timestamp into discrete components: \textit{year}, \textit{month}, \textit{day}, and \textit{hour}. Each component is mapped to a 16-dimensional embedding vector using separate embedding layers, each initialized randomly. Specifically, we use embedding layers with sizes: 7 for year (representing six possible years plus an index for unknown), 13 for month, 32 for day, and 25 for hour, to accommodate all possible values and account for unknown timestamps. The embeddings for the time components are concatenated to form a 64-dimensional time embedding for each timestep. 

Formally, given a batch size $B$, number of timesteps $t$, number of sources $s$, and number of patches $p$, we process the timestep tensor of shape $(B,\; t,\; 4)$, where the last dimension corresponds to [year, month, day, hour]. We apply the respective embedding layers to each time component, obtaining embeddings of shape $(B,\; t,\; 16)$ for each. These are concatenated along the last dimension to form the time embedding tensor of shape $(B,\; t,\; 64)$. The time embeddings are then expanded and combined with the positional encodings and source embeddings.

The source embeddings are generated by applying an embedding layer to the source labels, resulting in a tensor of shape $(B,\; s,\; d)$ with $d = 64$. After expanding and aligning dimensions, the time embeddings, positional encodings $(B,\; p,\; D)$, and source embeddings are concatenated along the feature dimension to form the final encoding tensor of shape $(B,\; t,\; s,\; p,\; D')$, where $D' = D + d + 64$.

This comprehensive encoding allows the model to incorporate temporal, positional, and source information, enhancing its ability to process data with multiple sources and variable timesteps, which is essential for practical remote sensing tasks. Note that since downstream tasks may not always have time information, we simulate missing timestamps in the data by occasionally setting the timesteps to zero vectors, representing unknown time components. Specifically, with a probability of 10\%, we replace the actual timesteps with zeros during training:

%To make sense of this information we use a learnable time encoding that embeds timestamp of the image, a learnable source encoding that embeds to which source and band each patch belongs and a fixed sine-cosine positional encoding.}

\subsection{Training Paradigm}
Our training approach uses a task-based system, and diverse masking strategies, and includes temporal information. These components work together to improve the flexibility and effectiveness of our EarthMAE model. This comprehensive approach matches well with the challenges presented by the various sensor data, timesteps, and masking schemes that are typical in the self-supervised learning of remote sensing data.

The training approach makes the most of the unique mix of data in our dataset. This includes varied sensors and data types, such as multispectral data from Sentinel, hyperspectral data from NEON, RGB data from Satellogic, and specific bands like Sentinel-2 RGB and SCL used for segmentation tasks. The training paradigm can be divided into tasks where each task is essentially a subset of the sources available in the data (e.g., RGG, CHM, etc), this simplifies the handling and combination of data sources of different channels, bands, sizes, and resolutions. To manage this broad spectrum of data, we have set up a task-based training system. Here, we distribute tasks across multiple GPUs, with each GPU handling a specific task. This way, we can process different types of data and sensors in parallel, making the process more efficient.
\vspace{-2.5mm}
\paragraph{Loss} The training objective is the mean squared loss (MSE) between the model's reconstruction and the normalized pixel values on masked patches same as in the original MAE~\cite{he2022masked} paper.
\vspace{-2.5mm}
\paragraph{Masking} Simply performing random masking such as in the original MAE can lead to very easy reconstruction tasks where the model can learn to translate from one source to another or copy most of a patch from another time step. To ensure challenging reconstruction tasks with video sequences, \cite{tongvideomae} proposes tube masking where a given patch is masked across all time steps to prevent \emph{information leak} through time. We incorporated tube masking in our implementation.
% One of the notable aspects of our model is its use of flexible masking strategies. Instead of sticking to the traditional tube masking~\cite{tongvideomae}, where all timesteps are masked the same way, we've also incorporated random masking, where timesteps are masked in a more arbitrary manner. We've even included a combined approach that merges the two strategies. These alternative masking methods have proven to be beneficial, leading to better results across our varied datasets. Particularly, random and combined masking seems to enable the model to better understand the temporal patterns in the data, which in turn boosts the model's performance.

% We leverage different task
% We explore different masking 
% We incorporate timesteps
% We incorporate segmentation loss

\begin{figure*}[h!]
     \centering
     \includegraphics[width=\textwidth]{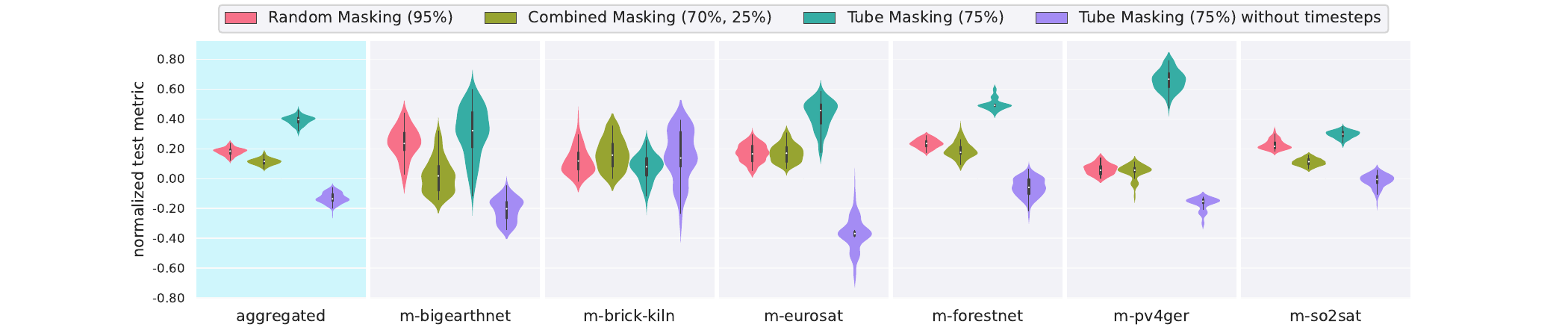}
     \caption{Performance of different masking schemes. Tube masking with a 75\% ratio consistently outperforms the rest. Combined masking refers to tube masking 75\% of the patches and randomly 25\% of the remaining ones. Pre-training without time information (purple) hurts performance. Results are reported across 5 different seeds.}
    \label{fig:time_mask_results}
\end{figure*}

\begin{figure*}[t!]
     \centering
     \includegraphics[width=\textwidth]{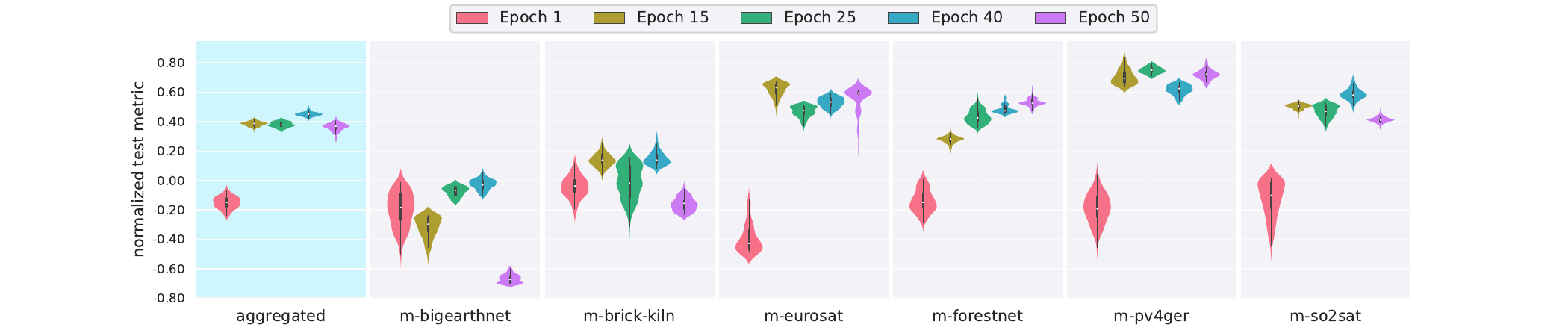}
     \caption{Performance on downstream tasks. Performance metrics are normalized and plotted for different epochs of model pre-training: Epochs 1, 15, 25, 40, and 50. The model shows varied performance across tasks, with some showing rapid improvement as early as Epoch 15. For example, m-forestnet and m-pv4ger exhibit a noticeable upward trend, suggesting that these tasks benefit early from the features learned during pre-training. Results are reported across 5 different seeds.}
    \label{fig:sat_progression}
\end{figure*}

Our experiments aimed to understand the impacts of different data sources (NEON, Sentinel, Satellogic), the inclusion of temporality, and various masking strategies on the performance of our Masked Autoencoder (MAE) model. 

\subsection{Experimental Setup}

To evaluate each pre-trained model, we leverage the classification benchmark of GeoBench \cite{lacoste2023geobench}. This benchmark is specifically designed to evaluate pre-trained models on remote sensing data. They curated 6 classification datasets: m-BigEarthNet, m-Brick-Kiln, m-EuroSAT, m-ForestNet, m-PV4GER, and m-SO2SAT, to cover a range of downstream tasks. On each downstream task, the pre-trained model is fine-tuned over parameters obtained through random search, and the best hyperparameter is selected on the validation set and re-trained with 5 different seeds to be evaluated on the test set. See \cref{sec:params} for training and fine-tuning details. A bootstrap procedure is used to report the uncertainty of the interquartile mean\footnote{Average discarding top and bottom 25\% to reduce outliers.}. A single aggregated result is obtained by averaging the normalized score\footnote{Normalization constants set weak baselines (e.g., ResNet18) to 0 and strong baselines (e.g., SwinV2) to 1.}.

% To fine-tune each experiment, we carried out an extensive hyper-parameter search, choosing the setup that yielded the highest validation accuracy. To ensure robust results, we ran each experiment with ten different seeds, reporting the test accuracy (or the F1-score for the BigEarthNet dataset) at the epoch with the best validation performance.

Following the standard MAE training process, all models were pre-trained for 100 epochs using a 90\% masking ratio with tube masking~\cite{tongvideomae}, where all timesteps and sources are masked in the same way (tube masking). We also experimented with different masking strategies by introducing random masking (where timesteps and sources are masked randomly with a 95\% ratio) and combined masking, which mixes tube and random masking strategies.

\subsection{Results and Discussion}
Our results showed significant variations in model performance, depending on the data sources, timesteps inclusion, and the applied masking schemes.

\cref{fig:time_mask_results} shows the performance for models trained on Sentinel data. Tube masking outperforms the rest in most downstream tasks while the removal of time information from the pre-training hurts performance in all datasets. For the results in this work, we use tube masking schema with a 75\% masking ratio, unless specified otherwise.

In \cref{fig:sat_progression}, we verify the convergence behavior of our training process, when training for 50 epochs with cosine annealing of the learning rate. The figure depicts the downstream performance for 5 checkpoints along the 50 epochs. The aggregated results show a mostly stable performance after 15 checkpoints, where the variability is likely due to the cross-checkpoint variance. Surprisingly, on certain tasks like m-bigearthnet, we observed an impressive drop in performance, that we were not able to explain. On the other hand, some tasks, like m-forestnet, expose a good progression of performance.

\begin{figure*}[t!]
     \centering
     \includegraphics[width=\textwidth]{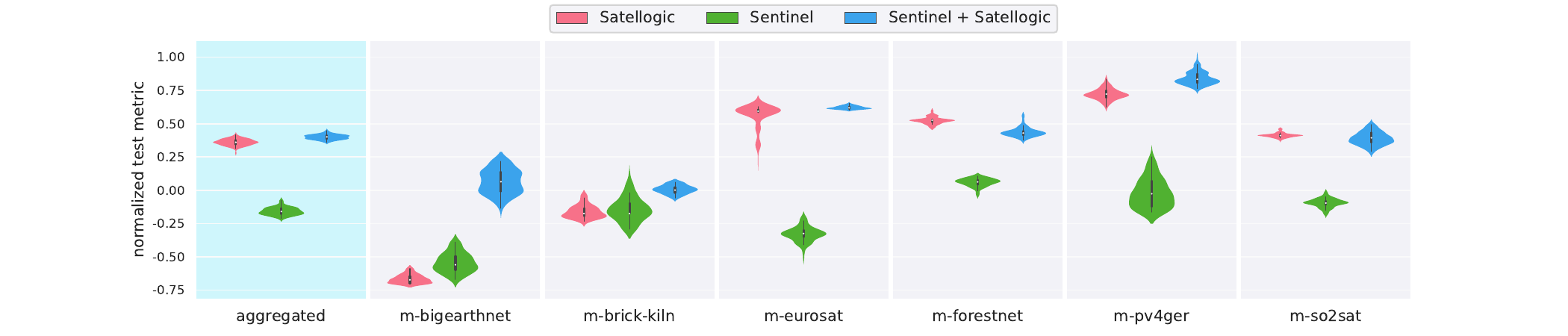}
     \caption{Downstream task performance for models pre-trained on different data. Note that including Satellogic data, whether alone or combined with Sentinel data, consistently enhances the model's performance across all tasks compared to using only Sentinel data. The combination of Sentinel and Satellogic data achieves the highest performance improvements. Results are reported across 5 different seeds.}
    \label{fig:normalized_metrics}
\end{figure*}

As seen in \cref{fig:normalized_metrics}, pre-training on Satellogic data offers consistent performance improvement over models pre-trained on Sentinel data alone, with the model that combines both data sources outperforming the rest. This suggests that the high resolution of Satellogic offers a useful signal during pre-training.

%Our findings highlight several key points. Firstly, Sentinel-2's (S2) RGB data consistently performs well across most datasets. However, adding other data types, specifically NEON's hyperspectral data and Sentinel-1's synthetic aperture radar (SAR) data, has varying effects on the model's performance. 

%Upon closer inspection, we noted a noticeable improvement in the average normalized test metric on datasets like m-BigEarthNet and m-Brick-Kiln when additional sensor data (NEON all and S1) were combined with the S2 RGB data. This indicates the potential benefits of using diverse sensor data, which could enhance the performance and reliability of remote sensing models. 

\begin{figure*}[t!]
     \centering
     \includegraphics[width=\textwidth]{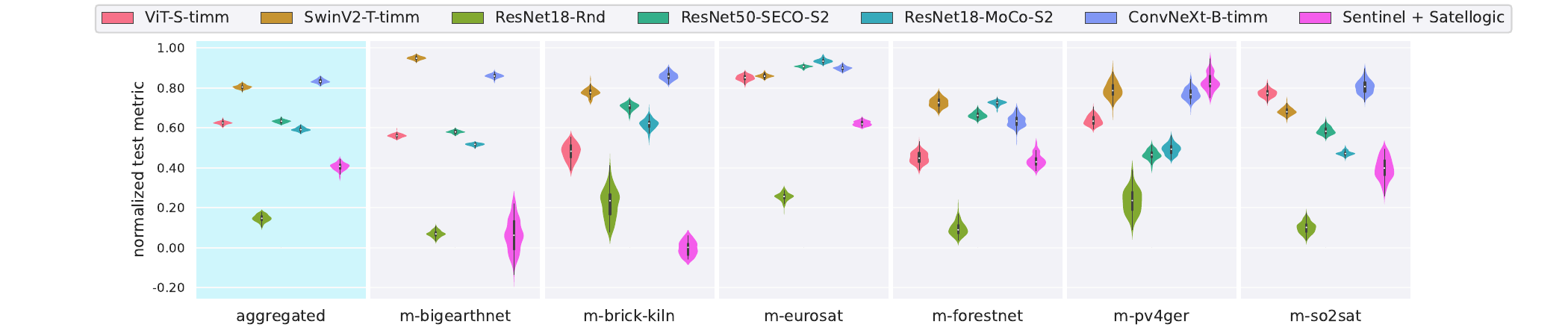}
     \caption{Downstream task performance of an MAE model pre-trained on Sentinel and Satellogic data and models featured on GeoBench \cite{lacoste2023geobench}.  Results are reported across 5 different seeds.}
    \label{fig:all_res}
\end{figure*}
\vspace{-2.5mm}

\section{Experiments}
Aggregated results in \cref{fig:all_res} show that the performances of this model are significantly lower than other pre-trained models evaluated in \cite{lacoste2023geobench}. Hypothesis for such discrepancies could be that MAE is not the right training loss or that remote sensing data alone is insufficient or too redundant.

\begin{figure}
     \centering
     \includegraphics[width=\columnwidth]{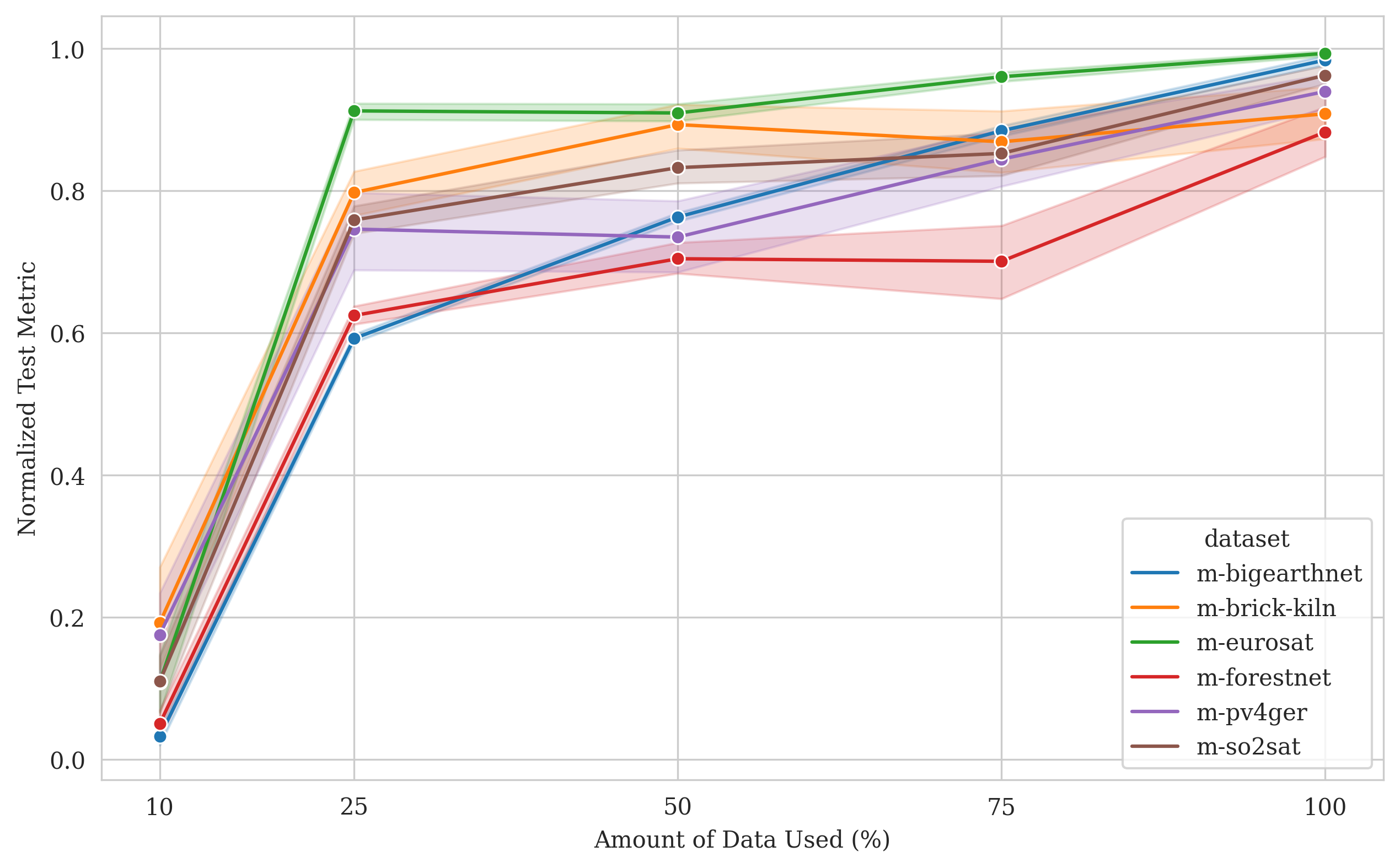}
     \setlength{\belowcaptionskip}{-23pt}
     \caption{Performance on downstream tasks for different dataset sizes shows that for some tasks, performance increases almost linearly, indicating considerable room for improvement with larger datasets.% Results are based on 10 seeds, with normalization done within these runs.
     }
    \label{fig:data_ratio_results}
    
\end{figure}

\cref{fig:data_ratio_results} shows how performance on downstream tasks varies for different dataset sizes. For harder tasks, the performance increases almost linearly indicating that despite the potential redundancy of satellite imagery, our dataset is diverse enough so that using it at full scale offers improved results. It seems that for tasks like m-bigearhtnet and m-forestnet would benefit from an even larger scale.

\section{Conclusion}

This paper primarily focuses on the introduction and description of our unique and expansive remote sensing dataset. With multiple sensors and data types, it provides researchers and practitioners with a wide array of information, which is anticipated to significantly advance the field of remote sensing and related applications.

While we have also presented EarthMAE, a tailored model designed to handle the nuances of our dataset, the overarching theme of our work is the sheer potential held within the dataset itself. The dataset’s size and diversity enable an exhaustive examination of different sensor types and data structures. Furthermore, the distribution of tasks across multiple GPUs during training fosters an efficient environment for exploring various self-supervised learning scenarios.

In our investigation, we explored different masking strategies each of which holds implications for the model's performance. We also incorporated temporal information from the dataset into the model using timestamp metadata, a strategy expected to increase accuracy across different remote sensing tasks.

Ultimately, the value of this dataset extends beyond the scope of our work. It provides an open playing field for future explorations in self-supervised learning and remote sensing applications. We hope that the efforts encapsulated in this paper serve as a springboard for future research.

\vspace{-3mm}
\paragraph{Limitations} While EarthView offers diverse sources, sensors, and scales, it lacks modalities like text. The EarthMAE model does not fully realize EarthView's potential. Researchers are encouraged to explore larger models trained on this dataset with others like \cite{bastani2022satlas}.  

% \bibliographystyle{tmlr}
% \bibliography{bibliography}

\bibliographystyle{ieee_fullname}
\bibliography{bibliography}

\newpage

\appendix
\newcommand{\tco}

\section{Impact of Fundamental Models on Earth Observation} 
\label{sec:impact}

Over the past few decades, remote sensing and Earth observation have had a transformative effect on a wide range of applications, including military, insurance, market prediction, and climate science, among others. Although this substantial impact cannot be directly ascribed to deep learning or large pre-trained networks, it forms part of a broader discussion that goes beyond the scope of this section. The focus here is on examining the role of fundamental models in enhancing Earth observation.

\subsection{Contributions to Climate Mitigation and Adaptation}
The use of machine learning on remote sensing data has become prevalent in devising solutions for an array of problems related to climate change \cite{burke2021using,rolnick2019tackling,zhu2017deep,ma2019deep}. These solutions are primarily designed by curating datasets for specific tasks, which necessitates considerable resources. In addition, these solutions are usually tailored to particular regions as extending the methods to new geographies continues to be a significant challenge, largely due to the scarcity of labelled data \cite{zhu2017deep}. Regions with less economic development, while equally vulnerable to the effects of climate change, often suffer from a deficit of effective remote sensing-based solutions \cite{burke2021using}. Fundamental models for Earth observation can potentially tackle many of these concerns, thereby significantly hastening and facilitating the creation of novel remote sensing solutions for climate change.

\subsection{Promoting Accessibility}
By diminishing the requirement to curate a large labelled dataset for each individual task, we could democratize the development of machine learning models for remote sensing, particularly for groups or entities operating on limited budgets \cite{maskey2020advancing, Alemohammad2021}. Fundamental models could be particularly beneficial for non-profit organizations, academic institutions, startups, and developing countries. They could also pave the way for applications that were previously not profitable. We posit that the wider availability of these models will primarily have a net positive impact, although we recognize that this access could lead to unforeseen applications with potentially negative effects \cite{bommasani2021opportunities}. Moreover, it's important to note that these models may have dual-use implications, where they could, for instance, aid oil and gas industries in their operations in a way that either increases or decreases overall emissions.

\subsection{Emissions from Large Pre-trained Models}
Recent studies have examined the emissions of large neural networks \cite{strubell2019energy, schwartz2020green, codecarbon, lacoste2019quantifying, patterson2021carbon}. Notably, training a large transformer can result in the emission of 284 \tco when run on computers primarily powered by fossil fuel energy (US national average) \cite{strubell2019energy}. When juxtaposed with individual actions, such emissions are substantial - a round-trip passenger flight from San Francisco to London results in 2.8 \tco, which is roughly 100 times smaller. Yet, the wide applicability of pre-trained models and their potential in aiding efforts to mitigate climate change \cite{rolnick2019tackling} prompts a shift in perspective.

Assessing new tools and systems necessitates a consideration of the probable net impact on emissions, both in terms of the tool's creation and its eventual deployment. For instance, testing the performance of airborne methane sensing tools at emission levels typically found in oil and gas operations can lead to the emission of about 7 metric tonnes of methane, roughly equivalent to 600 \tco over a 20-year global warming potential \cite{epa_greenhouse_2017}. Nevertheless, in a single day of operation, such an instrument can survey hundreds of sites, often identifying leaks that require repair and which emit considerably more than 7 metric tonnes of methane per day \cite{johnson_airborne_2021}. Similarly, fundamental models could significantly advance our capacity to utilize large amounts of passively collected satellite data, leading to massive reductions in emissions, enhancing our understanding of climate science qualitatively, and bolstering our ability to adapt to climate change.

In summary, the potential advantages for climate change mitigation through improved Earth observation methodologies likely outweigh the emissions associated with fundamental models. Furthermore, several actions can be undertaken to reduce and mitigate emissions linked to the training of your model \cite{lacoste2019quantifying}:
\begin{itemize}
\item Choose data centers that are certified as carbon neutral or predominantly powered by renewable energy, with efficient power usage (PUE). Such steps can drastically reduce emissions by about 50 times \cite{lacoste2019quantifying}.
\item Configure your code development process to minimize the need for computationally-intensive runs, for example, by using modular development and testing when possible.
\item Improve the efficiency of your code and sparsify your network where feasible \cite{patterson2021carbon}. This could reduce emissions by up to tenfold.
\item Opt for more energy-efficient hardware, such as TPUs or GPUs.
\item Monitor \cite{codecarbon} and report your emissions \cite{lacoste2019quantifying}. Better communication about climate change is vital for systemic changes. Improved documentation will assist other developers to continue from where you left off, possibly avoiding some computationally intensive runs.
\item Offset the cumulative emissions of your projects.
\end{itemize}

\subsection{Fairness and Biases}
It's well known that large language models can amplify and perpetuate biases \cite{bender2021dangers}. While this can lead to serious societal problems, we believe that biases in remote sensing models are likely to have a considerably lesser impact. However, we do foresee potential biases and fairness issues.

\paragraph{Data Coverage and Resolution} Certain satellites provide standard spatial resolution and revisit rate coverage for the entire Earth (e.g., Sentinel-2 offers global coverage at a resolution of 10-60 m/pixel every five days). This ensures that imagery is freely and uniformly available across the planet. Other satellite data providers, such as Maxar, provide images on demand and have a higher spatial resolution (up to 0.3m per pixel), but have lower revisit rates and higher costs. Some countries, such as New Zealand, freely offer aerial imagery with a resolution of up to 0.1m per pixel\footnote{\url{https://data.linz.govt.nz/}}. Finally, it's worth mentioning that cloudy seasons in certain climates may limit data availability for some countries. Overall, while coverage is relatively uniform, some regions have much higher coverage than others, and financial constraints can limit access to data. This can lead to some degree of biases and fairness issues.

\section{Hyper-parameters}
\label{sec:params}
The training and fine-tuning in our experiments follows the original MAE \cite{he2022masked} training paradigm.

All models were pre-trained using the same hyper-parameters:
\vspace{0.1cm}
\begin{itemize}
    \item \textbf{Effective batch size}: 2048 (32 per GPU $\times$ 64 GPUs)
    \item \textbf{Base learning rate (blr)}: $1.32 \times 10^{-4}$
    \item \textbf{Gradient clipping norm}: 1
    \item \textbf{Number of epochs}: 100
    \item \textbf{Warmup epochs}: 10
    \item \textbf{Weight decay}: 0.0457
\end{itemize}
\vspace{.5cm}
For fine-tuning, we use an effective batch size of 32 with a weight decay of $0.05$ and 5 warmup epochs.

The specific base learning rates (blr) used for each task were found through random search, see \ref{tab:finetuning_blr}.

\begin{table}[h!]
\centering
\caption{Base Learning Rates (blr) for Fine-tuning Tasks Using Different Models}
\label{tab:finetuning_blr}
\resizebox{\columnwidth}{!}{
\begin{tabular}{l c c c}
\hline
\textbf{Task} & \textbf{Sentinel} & \textbf{Satellogic} & \textbf{Sentinel + Satellogic} \\
\hline
m-bigearthnet & $9.76 \times 10^{-4}$ & $3.12 \times 10^{-4}$ & $7.17 \times 10^{-4}$ \\
m-so2sat & $6.53 \times 10^{-4}$ & $2.12 \times 10^{-4}$ & $4.57 \times 10^{-4}$ \\
m-brick-kiln & $1.86 \times 10^{-4}$ & $1.69 \times 10^{-5}$ & $1.81 \times 10^{-4}$ \\
m-forestnet & $5.09 \times 10^{-4}$ & $5.64 \times 10^{-4}$ & $1.81 \times 10^{-4}$ \\
m-eurosat & $7.57 \times 10^{-4}$ & $4.60 \times 10^{-4}$ & $1.65 \times 10^{-4}$ \\
m-pv4ger & $5.56 \times 10^{-4}$ & $2.72 \times 10^{-4}$ & $4.05 \times 10^{-4}$ \\
\hline
\end{tabular}
}
\end{table}

\begin{table}[ht]
    \centering
    \caption{Comparison of reconstruction losses under different training datasets and masking schemes. Satellogic data is generally harder to reconstruct due to its higher resolution. Random masking tends to be easier for the model to reconstruct, as it can leverage different bands and timesteps to recover missing information.}
    \resizebox{\columnwidth}{!}{
    \begin{tabular}{l l l}
    \hline
    \textbf{Training Data} & \textbf{Masking Schema} & \textbf{Reconstruction Loss} \\
    \hline
    Satellogic             & tunnel                  & 0.561          \\
    Satellogic             & random                  & 0.458           \\
    Sentinel               & tunnel                  & 0.285          \\
    Sentinel               & random                  & 0.284           \\
    Sentinel + Satellogic  & tunnel                  & 0.284           \\
    \hline
    \end{tabular}
    }
    \label{tab:reconstruction_losses}
\end{table}

\end{document}